\documentclass[11pt]{article}
\pdfoutput=1

\usepackage[preprint]{acl}

\usepackage{times}
\usepackage{latexsym}

\usepackage[T1]{fontenc}

\usepackage[utf8]{inputenc}

\usepackage{microtype}

\usepackage{inconsolata}

\usepackage{graphicx}
\usepackage{booktabs}
\usepackage{amsmath}

\title{Prompt-Robust Language Models: Which Training Strategies Work?}
\author{Frederic Sadrieh$^{1,2}$ \and Michal Štefánik$^{3}$ \vspace{6pt}\\
  $^1$\normalsize{Center for Information and Language Processing, LMU Munich} \\
  $^2$\normalsize{Munich Center for Machine Learning (MCML)} \\
  $^3$\normalsize{R\&D Centre for Large Language Models, National Institute of Informatics, Japan\vspace{6pt}} \\
  \small{Correspondence to \texttt{frederic.sadrieh@lmu.de}}
}

\begin{document}
\maketitle
\begin{abstract}
Despite their strong performance, large language models remain highly sensitive to prompt formulation. Prior work addresses this through refined data construction or through dedicated robustness objectives. We reproduce and compare these strategies under controlled conditions, and measure how effective they are in addressing models' prompt sensitivity. We find the current robustness fine-tuning methods improve over standard fine-tuning and in-context learning, but the best-to-worst prompt gap remains as high as 40–57\% of performance. Moreover, the recent robustness-enhancing methods we test --- \textsc{CoIN} for contrastive alignment and \textsc{PPCL} for consistency regularization --- often fail to outperform the simplest data construction strategy: training on one template per batch. Our diagnostics explain these results. The auxiliary objectives move the quantity they penalize, but do not \textit{generalize}. Additionally, data construction strategies differ due to the conflicting signs of per-template gradients on 57--64\% of parameters. Thus, batches that mix formulations force the optimizer to reconcile competing updates instead of finding a shared, prompt-agnostic one.\footnote{Our experiments can be reproduced with: \url{https://github.com/FSadrieh/prompt-agnostic-llms}}

\end{abstract}

\section{Introduction}
Large language models (LLMs) are increasingly deployed in real-world decision-making pipelines, yet their performance remains highly sensitive to prompt formulation \citep{inger-etal-2025-forget}. Semantically equivalent prompts can yield drastically different outputs, making systems brittle and unreliable in practice \citep{cao2024ontheworst}. Addressing this at inference time through manual prompt engineering is labor-intensive and fundamentally does not scale \citep{schulhoff2025promptreportsystematicsurvey}.

In contrast, train-time strategies build invariance directly into the model rather than patching sensitivity post-hoc. Prior work has shown that tuning on multiple prompt formulations already outperforms standard single-prompt instruction finetuning \citep{wei-etal-2025-paft, wei2022finetuned}, yet the design space remains poorly understood — it is unclear which data construction strategies matter, and whether more sophisticated robustness-enhancing methods provide meaningful gains on top.

We address this gap through a systematic study of train-time robustness methods improving on instruction finetuning (IFT)~\citep{wei2022finetuned}. We analyze IFT methods from three categories: (i)~data construction strategies mixing different prompt formulations \cite{wei-etal-2025-paft}, (ii)~consistency regularization methods \cite{qiang-etal-2024-ppcl}, and (iii)~contrastive methods \cite{yan-etal-2024-contrastive}. Our main contributions can be summarized as:
\begin{itemize}
\item We compare data construction strategies for multi-template IFT, and trace template interference in the gradient space, where per-template updates conflict in sign on 57--64\% of parameters;
\item We reproduce \textsc{CoIN} and \textsc{PPCL} and find both fail to reliably improve upon data construction strategies. Diagnostics of their loss objectives locate the cause in a failure to generalize;
\item We scope the limits of current train-time robustness methods: a best-to-worst prompt gap of 40--57\% survives every method we test.
\end{itemize}

Our work intends to be used as a guidance for engineers and researchers deciding which methods to prioritize when optimizing for prompt robustness.

\section{Related Work}
A large body of recent work documents that LLM performance remains highly sensitive to prompt formulation, regardless of model size and tasks \citep{Habba_2025, sun2024evaluating, zhu2024promptbench, sclar2024quantifying, mizrahi-etal-2024-state}. Prompt paraphrases or minor structural changes cause substantial performance swings \citep{chatterjee-etal-2024-posix, he2024doespromptformattingimpact, zhuo-etal-2024-prosa}. Critically, IFT raises mean performance but leaves prompt sensitivity largely intact \citep{chatterjee-etal-2024-posix, wei-etal-2025-paft}.

\textbf{Inference-time methods} address brittleness through prompt optimization, model calibration, or model-based self-correction \citep{zhao2021calibrate, hu2024localized, shi2025autopromptgenerationrobustprompt, zhan-etal-2024-unveiling}. While effective, these approaches add algorithmic overhead and do not generalize, requiring re-running at test time \citep{hu2024localized, shi2025autopromptgenerationrobustprompt}. \textbf{Train-time methods} instead build invariance directly into the model. Previous work experiments with multi-prompt IFT \citep{wei-etal-2025-paft} and adding explicit robustness objectives such as consistency regularization \citep{qiang-etal-2024-ppcl} and contrastive alignment \citep{yan-etal-2024-contrastive} on top. However, these studies do not provide a systematic comparison of methods from different families under controlled conditions.

Two recent studies are closest to ours: \citet{seleznyov-etal-2025-punctuation} evaluate \textsc{PPCL} and multi-prompt IFT and \citet{agrawal2025enhancing} include one consistency regularization method \citep{sun2024evaluating}, but both test local perturbations (e.g. separators) and primarily focus on test-time approaches. Our study is the first to compare robustness training methods from three categories under major prompt perturbations.

\section{Experimental setup}

\paragraph{Problem definition}
\label{sec:problem_def}

Given a collection of datasets $\{\mathcal{D}_1, \dots, \mathcal{D}_n\}$, each containing samples
$\mathcal{D}_{i} = \{(X_{i,1}, Y_{i,1}), \dots, (X_{i,m_i}, Y_{i,m_i})\}$ and associated with a set of
verbalization templates $\mathcal{T}_i = \{\mathcal{T}_{i,1}, \dots, \mathcal{T}_{i,k_i}\}$, where each
template maps a sample to a textual instruction and expected response,
$\mathcal{T}_{i,l}: (X_{i,j}, Y_{i,j}) \mapsto (x_{i,j,l}, y_{i,j,l})$.
We split the collection into a train $\mathcal{D}_\text{tr}$ and a test $\mathcal{D}_\text{te}$ part with
disjoint tasks. Applying every template of $\mathcal{T}_i$ to every sample of every
$\mathcal{D}_i \in \mathcal{D}_\text{tr}$ yields the instruction fine-tuning corpus
$\mathcal{X}_\text{IFT} = \{(x_{i,j,l}, y_{i,j,l}) \mid \mathcal{D}_i \in \mathcal{D}_\text{tr},\,
j \leq m_i,\, l \leq k_i\}$.
During training, we iteratively update the parameters $\theta$ of the model $\mathcal{M}_\theta$ on batches
$\mathcal{B} \subset \mathcal{X}_\text{IFT}$ as
$\theta \leftarrow \theta - \eta \nabla_\theta \mathcal{L}(\mathcal{B}; \theta)$.
Previous work differs in which $(i, j, l)$ triples are grouped into $\mathcal{B}$, or in which auxiliary
terms are added to $\mathcal{L}$.

Our experiments compare the effectiveness of these methods in achieving prompt robustness in $\mathcal{M}$. We employ each training method with $\mathcal{D}_\text{tr}$ and evaluate its resulting model on $\mathcal{D}_\text{te}$, for each $\mathcal{D}_i \in \mathcal{D}_\text{te}$ reporting performance with (i)~the best-performing template $\mathcal{T}_{i,\text{best}}$, (ii)~the worst-performing template $\mathcal{T}_{i,\text{worst}}$, and (iii)~the average performance across all templates $\mathcal{T}_{i}$.

\subsection{Training Methods}
\label{methods}

We study three categories of train-time methods to improve prompt robustness, all making use of the availability of multiple templates for each dataset. 

\paragraph{(1) Data construction}
We test four strategies for how batches are constructed:
\begin{itemize}
\item \textsc{Single} — Train on one template per dataset mirroring standard IFT \citep{wei2022finetuned}
\item \textsc{All shuffled} — All templates and examples are shuffled randomly into batches
\item \textsc{All-in-one-Batch} — All templates for a given example are used in the same batch
\item \textsc{One-at-a-Time} — Only one template is used per batch, mirroring \citet{wei-etal-2025-paft}.
\end{itemize}
With these strategies we test how varying prompt diversity per batch affects the model's robustness and assess the optimality of strategies applied in previous work.

\paragraph{(2) Consistency regularization — \textsc{PPCL}}
Consistency regularization methods add an auxiliary loss that penalizes divergence between the model's output probabilities on semantically equivalent prompts. We select \textsc{PPCL} \citep{qiang-etal-2024-ppcl} to represent this category, as it operates directly on the supervised finetuning objective and does not introduce other changes such as soft prompts \citep{sun2024evaluating} or unsupervised label construction \citep{zhou-etal-2022-prompt, hejabi-etal-2026-flip}. \textsc{PPCL} augments the standard cross-entropy loss, calculated for both prompt formulations, with a Jensen--Shannon divergence term computed over the average token probabilities of outputs across prompt formulations:
\begin{equation}
    \mathcal{L} = \lambda_1 \mathcal{L}_{CE\_orig} + \lambda_2 \mathcal{L}_{CE\_par} + \lambda_3 \cdot \mathcal{L}_{JSD}
\end{equation}
where $\lambda$-s control the trade-off between instruction following and prompt robustness.

\paragraph{(3) Contrastive alignment — \textsc{CoIN}}
Contrastive approaches align semantically equivalent prompts while separating those that are syntactically similar but semantically different. We select \textsc{CoIN} \citep{yan-etal-2024-contrastive} as it forms the basis of several subsequent approaches \citep{liuCREMERobustnessEnhancement2025, aissi-etal-2025-reinforcement}. \textsc{CoIN} adds a contrastive loss over the model's internal representations:
\begin{equation}
    \mathcal{L} = \mathcal{L}_{CE} + \lambda \cdot \mathcal{L}_{CoIN},
\end{equation}
where $\mathcal{L}_{CoIN}$ is the contrastive objective as defined in \citet{yan-etal-2024-contrastive}.

\subsection{Evaluation}
\label{eval}

\paragraph{Models} We evaluate all methods on four models spanning two model families and two scales: Llama3.2-1B and Llama3.1-8B \citep{grattafiori2024llama3herdmodels} and Qwen3-0.6B and Qwen3-8B \citep{yang2025qwen3technicalreport}. Using two families allows us to assess whether findings generalize across architectures, while the two scales test whether robustness effects are consistent across model capacity. For all models we use the base variants, ensuring that observed robustness differences are attributable to our training methods rather than prior instruction tuning. For the 8B models we apply LoRA \citep{hu2022lora} to reduce computational cost. For each model and loss function we sweep a separate learning rate; full hyperparameter details are in \autoref{app:hyp}.

\paragraph{Datasets} We replicate the data setup from \citet{sanh2022t0} testing for strict generalization by training on 48 datasets and testing on 11 datasets from unseen tasks. We use the PromptSource collection for our templates \citep{bach-etal-2022-promptsource}. It covers paraphrasing and structural changes for all our datasets — the perturbation types models are most sensitive to \citep{chatterjee-etal-2024-posix}. We manually check all templates and filter templates which are inapplicable or change the task type. We detail the template structure and show examples in \autoref{app:templates}. Datasets with only a single remaining template were removed, and each remaining dataset is capped at 10,240 training examples to balance contributions across tasks. For the full details on the datasets see \autoref{app:dataset}.
\textsc{CoIN} and \textsc{PPCL} require all prompt formulations to share the same label, so we sample the largest subset of templates whose expected labels agree. We call this the majority templates. \textsc{CoIN} requires at least three majority templates per dataset, so we exclude all datasets with less than 3 such templates from majority training. To measure how much this affects performance, we also train the \textsc{All shuffled} method using only the majority templates.

\paragraph{Metrics} Following \citet{wang-etal-2022-super}, we use Rouge-L \citep{lin-2004-rouge} as our main metric. Each method is assessed on three metrics relative to IFT: average, worst-case and best-case template performance. In addition, we report rank classification accuracy, the original metric of the dataset from \citet{sanh2022t0}, in \autoref{app:rankcls}.
We rank all runs by score, take the top run as reference, and run a paired one-sided bootstrap test of reference-vs-competitor for every other run. Runs denoted as \textit{statistically best} are the ones that are not significantly worse than the best-performing run ($\alpha = 0.05$).

\section{Results}

\begin{figure*}[ht]
    \centering
    \includegraphics[width=\textwidth]{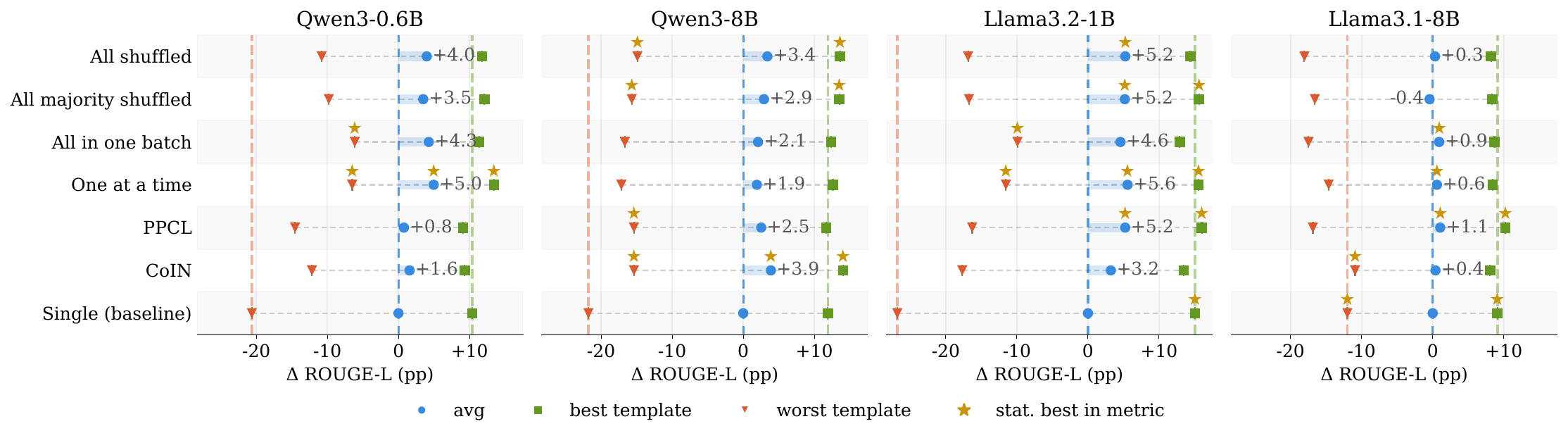}
    \caption{Performance of each method relative to IFT in percentage points across models. The circle, square, and triangle mark the average, best, and worst template score respectively; the dashed span reflects sensitivity to prompt variance. Vertical lines extending from IFT serve as reference. A star above a marker indicates membership in the statistically best group. The \textsc{Single} reference points are avg Rouge-L of 0.514 for Qwen3-0.6B, 0.634 for Qwen3-8B, 0.479 for Llama3.2-1B and 0.639 for Llama3.1-8B. See \autoref{tab:rougeL_raw} for all raw scores.}
    \label{fig:per_model_robustness}
\end{figure*}

\paragraph{Impact of robustness training}
First we evaluate the impact of robustness training methods. In \autoref{tab:rougeL_raw} we compare (i)~the untuned base models, (ii)~four-shot in-context learning (\textsc{ICL}), and (iii)~the robustness-trained models. Compared to the base model, IFT achieves much higher performance (an average Rouge-L score of 0.479 vs. 0.051 on Llama3.2-1B). ICL improves upon the base model (0.192 on Llama3.2-1B), but on three of the four it fails to reach IFT's performance level. The untuned Llama models in particular largely fail to follow the prompt format, so additional training is crucial. The exception is Qwen3-8B, whose base model is by far the strongest (0.434 average Rouge-L against 0.051--0.093). Here \textsc{ICL} surpasses IFT on average (0.660 vs. 0.634), but it still falls below the best multi-prompt IFT methods on average (0.673) and on the worst template (0.469 vs. 0.477--0.485). \textsc{ICL} is thus only competitive when the base model is already strong, and even then it does not win on robustness. The two strategies also differ in cost structure: robustness training is a one-time cost, whereas in-context learning adds a recurring cost to every inference call, so the choice depends on the strength of the base model and on the expected inference volume.

Across models, the evaluated multi-prompt IFT methods generally improve across all metrics relative to the IFT baseline (see \autoref{fig:robustness}), showing the importance of a high prompt variance in training. To rule out that these gains stem solely from the additional training steps, we also train a step-matched IFT run for five epochs. \autoref{tab:rougeL_raw} shows that it still underperforms. Still, all gains from multi-prompt IFT are modest relative to the remaining performance variance, which reaches 40--57\% across methods (see \autoref{fig:robustness}).

\paragraph{Impact of model choice}
\autoref{fig:per_model_robustness} reveals that smaller models are disproportionately sensitive to prompt formulation. For Llama3.2-1B, IFT yields a worst-to-best spread of 0.419 Rouge-L, indicating substantial instability. The larger models present a more nuanced picture: while Qwen3-8B exhibits a comparable variance span (0.337 vs.\ 0.310 Rouge-L), Llama3.1-8B displays considerably lower sensitivity to prompt formulation. On this model, several multi-prompt IFT methods instead produce statistically significant \emph{decreases} in worst-case performance. We attribute this to two factors: (1) the model's strong IFT performance reduces the ceiling for robustness gains, and (2) the validation performance improves mainly within the first 1{,}000 steps, with subsequent gains small and inconsistent (\autoref{app:valcurves}).

\paragraph{RQ1: How does data construction affect robustness?}
Prior work constructs multi-prompt IFT-data using a sequential per-template schedule \citep{wei-etal-2025-paft}, but it is unclear whether this specific construction is necessary or optimal. \autoref{fig:per_model_robustness} shows that for smaller models, \textsc{One-at-a-Time} yields consistent improvements, achieving statistically best or equivalent-to-best performance across all metrics. The worst-template gains are the largest: from $0.210$ to $0.363$ Rouge-L on Llama3.2-1B and from $0.308$ to $0.449$ on Qwen3-0.6B. \textsc{All-in-One-Batch} narrows the performance spread across formulations compared to \textsc{One-at-a-Time}, but it does so primarily by degrading best-case performance rather than lifting the worst case. Putting all templates of an example into the same update step, as done in \textsc{All-in-One-Batch}, leads to \textbf{interfering gradients} rather than the emergence of one prompt-agnostic shared update. We investigate this effect by separating the individual template gradients and measuring their interference. The resulting update vectors have a mean pairwise cosine similarity of 0.54 and a sign conflict (fraction of parameters in which template updates have conflicting signs) on 57--64\% of weights. This interference is consistent throughout training and strongest in the earliest blocks, which are most dependent on the prompt formulation (\autoref{app:interference}). In contrast, the methods with template-homogeneous batches never co-locate these conflicting gradients in one update and thus perform better. \textsc{All shuffled} and \textsc{All-Majority-Shuffled} perform comparably, since restricting
training to the majority templates discards only a small fraction of the templates per dataset
(\autoref{app:dataset}).

\begin{figure*}[htbp]
    \centering
    \includegraphics[width=\textwidth]{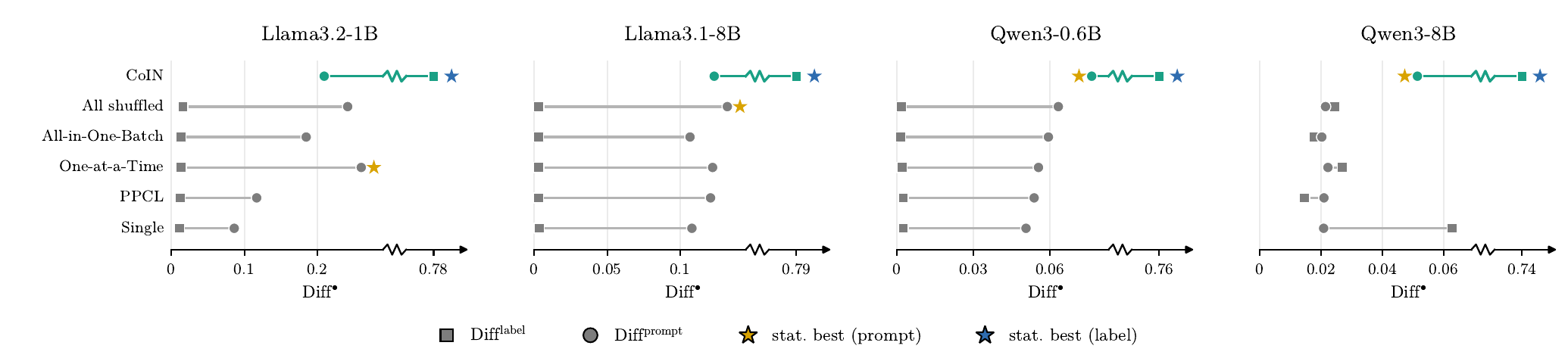}
    \caption{The hidden-state difference on the test split. Squares show the difference at the \texttt{label} readout, while circles show the difference at the \texttt{prompt} readout. A star next to a marker indicates membership in the statistically best group for that readout.}
    \label{fig:margin_per_model}
\end{figure*}
\paragraph{RQ2: Is the additional complexity of \textsc{PPCL} and \textsc{CoIN} justified?}

\textsc{PPCL} and \textsc{CoIN} both yield only modest gains compared to the baseline (\autoref{fig:robustness}). These gains are consistently matched or exceeded by simpler data construction strategies (\autoref{fig:per_model_robustness}). The sole exception is \textsc{CoIN} on Qwen3-8B, where it achieves the highest average performance of all evaluated methods; on all remaining models, \textsc{CoIN} and \textsc{PPCL} are either statistically indistinguishable from or inferior to the best data construction approach. The rank classification accuracy results mostly agree with these findings (see \autoref{app:rankcls}). The exception is Llama3.1-8B, where \textsc{PPCL} attains the best average and worst-template accuracy. It is also the only model whose multi-template methods do not consistently outperform single-template training on worst-template Rouge-L.

\paragraph{Why do \textsc{PPCL} and \textsc{CoIN} fail to improve?}

We analyze whether the two auxiliary objectives achieve their intended effect: grouping semantically similar prompts in the hidden states for \textsc{CoIN}, and matching output distributions across templates for \textsc{PPCL}. We measure each at two positions: (1)~the \texttt{label} readout over the label tokens, where the objectives are applied in training, and (2)~the \texttt{prompt} readout at the last prompt token (\autoref{app:diagnostics}).

\textsc{CoIN} achieves the largest contrastive difference among all methods at the \texttt{label} readout: $0.74$--$0.79$ against at most $0.063$ for others (see \autoref{fig:margin_per_model}). Since the \texttt{label} readout is used in training, the objective shifts its target as intended, while all other methods stay near zero, consistently with findings of \citet{aissi-etal-2025-reinforcement}. However, at the \texttt{prompt} readout, this advantage shrinks to at most $0.03$ on the Qwen models, and for the Llama models, \textsc{CoIN} separates prompts worse than the strongest data construction approaches. These findings demonstrate that \textsc{CoIN} can separate prompts when paired with their labels, but not the prompt alone.

\textsc{PPCL} attains the lowest cross-template JS-divergence of any method on the \emph{validation} split for both Llama models (see \autoref{fig:jsdiv_per_model} and \autoref{tab:ppcl_diag}), thus succeeds in-distribution. However, on the held-out test split the advantage disappears: \textsc{PPCL} falls to the worst multi-template method on Llama3.2-1B. For the Qwen models \textsc{PPCL} is not even the lowest on the validation split. Unlike \textsc{CoIN}, \textsc{PPCL} never opens a comparable gap over the other multi-prompt IFT methods (25--214$\times$ vs. at most 1.19$\times$). Every multi-template schedule already reduces divergence roughly two- to fivefold over single-template training in validation, so an explicit consistency term adds little on top of simply training on varied templates. On the two small models, \textsc{PPCL} consequently shows the largest validation-to-test gap of any method, so its distribution alignment does not survive the shift. Ultimately, we find that both \textbf{objectives improve in the training setting, but this improvement does not generalize}.

\section{Conclusion}

This paper compares train-time strategies for improving the prompt robustness of LLMs. All multi-prompt IFT methods improve over IFT, yet the gains stay modest relative to the underlying variance caused by prompt formulation. Among data construction strategies the differences are subtle, but \textsc{One-at-a-Time} data selection, where only samples of the same template share a batch and therefore an update, provides the most reliable worst-case gains. In contrast, the \textsc{All-in-One-Batch} strategy, which collects all templates of a sample in a single batch, underperforms due to gradient interference: conflicting signs across templates cancel each other out. More sophisticated methods from prior work, concretely \textsc{PPCL} and \textsc{CoIN}, are mostly matched or outperformed by simple data construction, making their additional complexity difficult to justify. We find that their additional loss objectives fail to generalize beyond the training setting. Therefore, we advise practitioners to prioritize simple \textsc{One-at-a-Time} or \textsc{All shuffled} schedules before investing in consistency regularization or contrastive objectives. 
Our results show that batch construction alone moves prompt robustness substantially, yet even the best train-time method we study leaves ample room for improvement.

\section*{Limitations}

\paragraph{Model scale and family coverage.}
Our experiments are limited to models up to 8B parameters across two model families (Llama and Qwen), due to computational constraints. It remains an open question whether the observed patterns — in particular the advantage of \textsc{One-at-a-Time} scheduling and the marginal benefit of auxiliary loss objectives — hold at larger scales where models may be more capable of leveraging richer training signals. Expanding the evaluation to additional architectures would further strengthen the generalizability of our conclusions.

\paragraph{Benchmark scope and evaluation protocol.}
We adopt the training and evaluation split of \citet{sanh2022t0}, which, while well-established, consists predominantly of classification and short-answer tasks with closed label sets. In such settings, prompt sensitivity may manifest differently than in open-ended generation, so evaluating on more diverse benchmarks — including long-form generation and reasoning tasks — would provide a more complete picture of prompt robustness.

\paragraph{Method selection within each category.}
We evaluate a single representative method for each of the consistency regularization and contrastive alignment categories — \textsc{PPCL} and \textsc{CoIN}, respectively — selected based on their central standing in those categories. Alternative methods within these categories \citep{sun2024evaluating, zhou-etal-2022-prompt, hejabi-etal-2026-flip, liuCREMERobustnessEnhancement2025, aissi-etal-2025-reinforcement} may exhibit different trade-offs, and our conclusions should be interpreted as applying to the specific instantiations studied rather than to the broader methodological families.

\paragraph{Hyperparameter optimization.}
We swept the learning rate separately for every model and every loss function, so \textsc{PPCL} and \textsc{CoIN} received the same per-model tuning budget as the data-construction baselines (\autoref{tab:lr}). The auxiliary loss weights, however, were not tuned per model. We swept the \textsc{PPCL} weight $\lambda_3$ on Llama3.2-1B and Qwen3-0.6B, selecting $\lambda_3 = 1$ on both, with performance being stable from 0.1 to 10 on Qwen3-0.6B (\autoref{tab:ppcl_lambda}). The \textsc{CoIN} hyperparameters $\tau$ and $\lambda$ were adopted from \citet{yan-etal-2024-contrastive} and not tuned at all. A full per-model sweep of every auxiliary-loss hyperparameter was prohibitive within our compute budget, and we cannot exclude that it would narrow the gaps we report. Our diagnostics (\autoref{app:diagnostics}) argue against under-tuning as the sole explanation, since both objectives measurably move the quantity they penalize.

\section*{Acknowledgments}
The project on which this report is based was funded by the Federal Ministry of Research, Technology and Space under the funding code “KI-Servicezentrum Berlin-Brandenburg” 16IS22092. Responsibility for the content of this publication remains with the author. We would like to thank Konstantin Dobler and Gerard de Melo for their continued support and feedback throughout this project.

\bibliography{references}

\appendix

\section{Dataset statistics}
\label{app:dataset}
\autoref{tab:train_datasets} shows an overview of all training datasets mirroring the T0 \citep{sanh2022t0} training split. We replaced tydiqa with Zaid/coqa\_expanded and gigaword with scitldr as the versions on Hugging Face were not compatible with most prompt templates in PromptSource. 

Afterward we manually filter out all templates which are not applicable or change the task type. For example, some templates reformulated HellaSwag examples as topic classification problems, which would cause train/test leakage.

After removing these templates we needed to remove social\_i\_qa, riddle\_sense, and wiki\_bio as they did not have more than one template remaining. Note all changes in dataset selection only affected the training split.

\begin{table*}[htbp]
  \centering
  \small
  \caption{
    The 48 training datasets grouped by task type. \emph{Templates} is the total number of PromptSource templates per dataset. \emph{Filtered} is the count after removing manually identified bad templates, and \emph{Majority} is the size of the majority templates (see \autoref{eval}). The number of samples is limited to 10,240 per dataset. We use the evaluation splits of the datasets for validation, with a sample limit of 512 and the same templates.
  }\label{tab:train_datasets}
  \begin{tabular}{@{}llrrrr@{}}
    \toprule
    \textbf{Task} & \textbf{Dataset} & \textbf{Samples} & \textbf{Templates} & \textbf{Filtered} & \textbf{Majority}\\
    \midrule
  Paraphrase        & GLUE (MRPC)           &  3,668 &  7 &  5 &  4 \\
                    & GLUE (QQP)            & 10,240 &  6 &  6 &  4 \\
                    & PAWS                  & 10,240 & 12 & 10 & 10 \\
  \midrule
  Extractive QA     & HotpotQA              & 10,240 &  5 &  5 &  5 \\
                    & TriviaQA              & 10,240 &  5 &  4 &  1 \\
                    & WebQuestions          &  3,778 &  5 &  5 &  1 \\
                    & WikiQA                & 10,240 & 11 &  5 &  4 \\
                    & AdvQA (BiDAF)         & 10,000 &  5 &  4 &  4 \\
                    & AdvQA (BERT)          & 10,000 &  5 &  4 &  4 \\
                    & AdvQA (RoBERTa)       & 10,000 &  5 &  4 &  4 \\
                    & DuoRC (Self)          & 10,240 &  9 &  5 &  1 \\
                    & DuoRC (Paraphrase)    & 10,240 &  9 &  5 &  1 \\
                    & ROPES                 & 10,240 & 12 & 12 & 12 \\
                    & SQuAD v2              & 10,240 & 12 &  4 &  2 \\
                    & SuperGLUE (ReCoRD)    & 10,240 & 20 &  7 &  1 \\
                    & Quoref                & 10,240 & 11 & 10 &  1 \\
                    & CoQA                  & 10,240 &  7 &  7 &  6 \\
  \midrule
  MC QA             & ARC-Challenge         &  1,119 &  6 &  5 &  3 \\
                    & ARC-Easy              &  2,251 &  6 &  5 &  3 \\
                    & CoS-E                 &  9,741 & 11 &  6 &  3 \\
                    & CosmosQA              & 10,240 & 13 &  8 &  4 \\
                    & DREAM                 &  6,116 &  5 &  4 &  2 \\
                    & OpenBookQA            &  4,957 &  7 &  7 &  6 \\
                    & PIQA                  & 10,240 & 11 &  6 &  3 \\
                    & QASC                  &  8,134 &  8 &  6 &  6 \\
                    & QuAIL                 & 10,240 & 13 & 11 &  6 \\
                    & QuaRel                &  1,941 &  5 &  5 &  5 \\
                    & QuaRTz                &  2,696 &  8 &  8 &  8 \\
                    & RACE (High)           & 10,240 &  8 &  5 &  3 \\
                    & RACE (Middle)         & 10,240 &  8 &  5 &  3 \\
                    & SciQ                  & 10,240 &  5 &  5 &  5 \\
                    & SuperGLUE (BoolQ)     &  9,427 & 10 &  5 &  3 \\
                    & SuperGLUE (MultiRC)   & 10,240 & 10 & 10 & 10 \\
                    & WIQA                  & 10,240 &  8 &  4 &  2 \\
  \midrule
  Sentiment         & Amazon Polarity       & 10,240 &  9 &  9 &  3 \\
                    & App Reviews           & 10,240 &  4 &  4 &  1 \\
                    & IMDB                  & 10,240 & 11 & 11 &  7 \\
                    & Rotten Tomatoes       &  8,530 & 10 & 10 &  7 \\
                    & Yelp Reviews          & 10,240 &  7 &  7 &  5 \\
  \midrule
  Summarization     & CommonGen             & 10,240 &  9 &  7 &  7 \\
                    & CNN/DailyMail         & 10,240 &  9 &  7 &  7 \\
                    & SciTLDR               &  1,992 &  6 &  5 &  5 \\
                    & MultiNews             & 10,236 &  6 &  5 &  5 \\
                    & SAMSum                & 10,240 &  7 &  6 &  5 \\
                    & XSum                  & 10,240 & 10 & 10 & 10 \\
  \midrule
  Topic Classification    & AG News               & 10,240 &  7 &  7 &  4 \\
                    & DBpedia               & 10,240 &  4 &  4 &  4 \\
                    & TREC                  &  5,452 &  18 &  5 &  5 \\
  \midrule
  \multicolumn{2}{@{}l}{\textbf{Total (48 datasets)}} & \textbf{417,238} & & & \\
    \bottomrule
  \end{tabular}
\end{table*}

\begin{table*}[htbp]
  \centering
  \caption{
    The 11 held-out test datasets grouped by task type. \emph{Templates} is the total number of PromptSource templates per dataset. \emph{Filtered} is the count after removing manually identified bad templates, and \emph{Majority} is the size of the majority templates (see \autoref{eval}).
  }\label{tab:test_datasets}
  \begin{tabular}{@{}llrrrr@{}}
    \toprule
    \textbf{Task} & \textbf{Dataset} & \textbf{Samples} & \textbf{Templates} & \textbf{Filtered} & \textbf{Majority}\\
    \midrule
    NLI               & SuperGLUE (CB)        &     56 & 15 & 15 &  8 \\
                      & SuperGLUE (RTE)       &    277 & 10 & 10 &  9 \\
                      & ANLI (R1)             &  1,000 & 15 & 15 &  8 \\
                      & ANLI (R2)             &  1,000 & 15 & 15 &  8 \\
                      & ANLI (R3)             &  1,200 & 15 & 15 &  8 \\
    \midrule
    WSD               & SuperGLUE (WiC)       &    638 & 10 & 10 &  9 \\
    \midrule
    Coreference       & SuperGLUE (WSC)       &    104 & 10 & 10 &  7 \\
    \midrule
    Commonsense       & WinoGrande            &  1,767 &  6 &  6 &  5 \\
                      & SuperGLUE (COPA)      &    100 & 12 &  8 &  8 \\
                      & StoryCloze            &  1,871 &  6 &  5 &  5 \\
                      & HellaSwag             & 10,003 & 11 &  5 &  3 \\
    \midrule
    \multicolumn{2}{@{}l}{\textbf{Total (11 datasets)}} & \textbf{18,016} & & & \\
    \bottomrule
  \end{tabular}
\end{table*}

\section{Template Structure and Examples}
\label{app:templates}

We use the PromptSource \citep{bach-etal-2022-promptsource} templates associated with each dataset without modifying their functionality. Each template converts an input example into an instruction, $x$, and the verbalization of the expected answer, $y$. We change the wording, framing, and answer-option ordering of the prompt through the templates, while keeping the underlying task fixed. The number of templates per dataset before and after filtering is given in \autoref{tab:train_datasets} and \autoref{tab:test_datasets}.

\autoref{fig:copa_templates} shows the eight filtered SuperGLUE~(COPA) templates rendered on the same example. The four templates that were removed failed to render with the current version of the dataset on Hugging Face, so they are not shown here. This example illustrates the magnitude of perturbation covered by our benchmark in rephrasing the question and modifying the prompt structure.

\begin{figure}[htbp]
  \centering
  \small
  \begin{quote}
    My view of the movie screen was blocked. What's the best option?\\
    -- The couple behind me was whispering\\
    -- A tall person was sitting in front of me\\
    We are looking for a cause
  \end{quote}
  \rule{\linewidth}{0.4pt}
  \begin{quote}
    My view of the movie screen was blocked. Select the most plausible cause:\\
    -- The couple behind me was whispering\\
    -- A tall person was sitting in front of me
  \end{quote}
  \rule{\linewidth}{0.4pt}
  \begin{quote}
    My view of the movie screen was blocked because... Choose between:\\
    --The couple behind me was whispering\\
    -- A tall person was sitting in front of me
  \end{quote}
  \rule{\linewidth}{0.4pt}
  \begin{quote}
    Exercise: choose the most plausible alternative. My view of the movie screen was
    blocked because...\\
    -- The couple behind me was whispering\\
    -- A tall person was sitting in front of me
  \end{quote}
  \rule{\linewidth}{0.4pt}
  \begin{quote}
    Pick the more likely continuation to the following sentence: My view of the movie screen
    was blocked as a result of:\\
    -- The couple behind me was whispering\\
    -- A tall person was sitting in front of me
  \end{quote}
  \rule{\linewidth}{0.4pt}
  \begin{quote}
    My view of the movie screen was blocked. This happened because... Help me pick the more
    plausible option:\\
    -- The couple behind me was whispering\\
    -- A tall person was sitting in front of me
  \end{quote}
  \rule{\linewidth}{0.4pt}
  \begin{quote}
    My view of the movie screen was blocked. I am hesitating between two options. Help me
    choose the more likely cause:\\
    -- The couple behind me was whispering\\
    -- A tall person was sitting in front of me
  \end{quote}
  \rule{\linewidth}{0.4pt}
  \begin{quote}
    ``The couple behind me was whispering'' or ``A tall person was sitting in front of
    me''? My view of the movie screen was blocked, because
  \end{quote}

  \caption{
    All filtered SuperGLUE~(COPA) templates rendered on one example.
  }\label{fig:copa_templates}
\end{figure}

\section{Hyperparameters}
\label{app:hyp}

\autoref{tab:shared_hyp} lists the hyperparameters shared across all models and methods. Due to compute constraints we only sweep for the first 2{,}000 training steps and take the best value from there.
\autoref{tab:lr} reports the learning rates selected per model and method via grid search over $\{10^{-5}, 3{\times}10^{-5}, 5{\times}10^{-5}, 7{\times}10^{-5}, 9{\times}10^{-5}, 10^{-4}\}$.

The $\lambda_3$ hyperparameter for \textsc{PPCL} was tuned over $\{0.1, 1, 10, 100\}$ on Llama3.2-1B. To check the robustness of this hyperparameter, we also tuned it on Qwen3-0.6B using the same protocol. \autoref{tab:ppcl_lambda} reports the Qwen3-0.6B sweep: $\lambda_3 = 1$ falls in the statistically best group on all three metrics, and performance is stable from $0.1$ to $10$, collapsing only at $100$. The \textsc{CoIN} hyperparameters $\tau$ and $\lambda$ were set to the advised values from \citet{yan-etal-2024-contrastive} and not tuned further due to computational constraints.

All experiments are run on H200 GPUs, and we trained for roughly 2{,}800 GPU hours.

\begin{table*}[htbp]
  \centering
  \caption{Shared hyperparameters for all experiments. $^\dagger$LoRA is only applied to the 8B models.}\label{tab:shared_hyp}
  \begin{tabular}{@{}ll@{}}
    \toprule
    \textbf{Hyperparameter} & \textbf{Value} \\
    \midrule
    Optimizer               & AdamW (fused) \\
    LR schedule             & Constant with linear warmup \\
    Warmup steps            & 100 \\
    Effective batch size    & 256 \\
    Epochs                  & 1 \\
    Precision               & bfloat16 \\
    Max train samples per dataset & 10\,240 \\
    \midrule
    LoRA rank$^\dagger$     & 16 \\
    LoRA alpha$^\dagger$    & 32 \\
    LoRA dropout$^\dagger$  & 0.0 \\
    LoRA target modules$^\dagger$ & \texttt{q\_proj}, \texttt{v\_proj}, \texttt{output\_proj}, MLP \\
    \midrule
    \textsc{CoIN} $\tau$             & 0.05 \\
    \textsc{CoIN} $\lambda$          & 1000 \\
    \textsc{PPCL} $\lambda_1$, $\lambda_2$ & 1, 1 \\
    \textsc{PPCL} $\lambda_3$        & 1 \\
    \bottomrule
  \end{tabular}
\end{table*}

\begin{table*}[htbp]
  \centering
  \caption{Learning rates selected per model and method.
    All template variants (single, all, etc.) share the same standard loss function, therefore we tune the learning rate only once.}\label{tab:lr}
  \begin{tabular}{@{}lccc@{}}
    \toprule
    \textbf{Model} & \textbf{Standard Loss} & \textbf{\textsc{CoIN}} & \textbf{\textsc{PPCL}} \\
    \midrule
    Llama-3.2-1B & $5{\times}10^{-5}$ & $7{\times}10^{-5}$ & $3{\times}10^{-5}$ \\
    Llama-3.1-8B & $5{\times}10^{-5}$ & $9{\times}10^{-5}$ & $10^{-4}$          \\
    Qwen3-0.6B   & $3{\times}10^{-5}$ & $5{\times}10^{-5}$ & $10^{-5}$          \\
    Qwen3-8B     & $10^{-5}$          & $5{\times}10^{-5}$ & $10^{-5}$          \\
    \bottomrule
  \end{tabular}
\end{table*}

\begin{table}[htbp]
  \centering
  \small
  \caption{
    \textsc{PPCL} loss-weight sweep on Qwen3-0.6B, Rouge-L on the test split.
    \textbf{Bold} values belong to the statistically best group per column.
  }\label{tab:ppcl_lambda}
  \begin{tabular}{@{}lccc@{}}
    \toprule
    \textbf{Run} & \textbf{avg} & \textbf{best} & \textbf{worst} \\
    \midrule
    $\lambda_3 = 0.1$   & \textbf{0.5087} & \textbf{0.6391} & \textbf{0.3487} \\
    $\lambda_3 = 1.0$   & \textbf{0.5088} & \textbf{0.6385} & \textbf{0.3587} \\
    $\lambda_3 = 10.0$  & 0.4945          & \textbf{0.6309} & \textbf{0.3670} \\
    $\lambda_3 = 100.0$ & 0.3528          & 0.4517          & 0.1099          \\
    \bottomrule
  \end{tabular}
\end{table}

\section{Additional Results}
\label{app:results}

\paragraph{Inference-time baselines}
The \textsc{Base model} and \textsc{ICL} rows of \autoref{tab:rougeL_raw} evaluate the untrained base checkpoints. No weights are loaded or updated in either case, so both isolate prompting from training and are scored exactly like every other test run.
For \textsc{ICL} we prepend four demonstrations to each query. The demonstrations are drawn from the evaluation split itself, excluding the query example, since the eleven test tasks are held out from our training mixture and therefore have no in-task examples in it. Each demonstration is rendered with the \emph{same} template as the query and inserted as a prompt/answer turn pair carrying the gold label, so the entire context is verbalised consistently.
We sample the four demonstrations once per example, using a seed derived from the dataset and example identifier, and reuse that draw for all templates of the example. Any variation in \textsc{ICL} across templates is therefore attributable to the template rather than to a different choice of demonstrations, which keeps the comparison with the trained models on equal footing.
Drawing demonstrations from the evaluation split supplies \textsc{ICL} with in-distribution, gold-labelled examples of the target task. This favours the inference-time baseline, which we nevertheless find to stay below the best multi-template method on every model and on every metric.

\begin{figure}[htbp]
    \centering
    \includegraphics[width=\columnwidth]{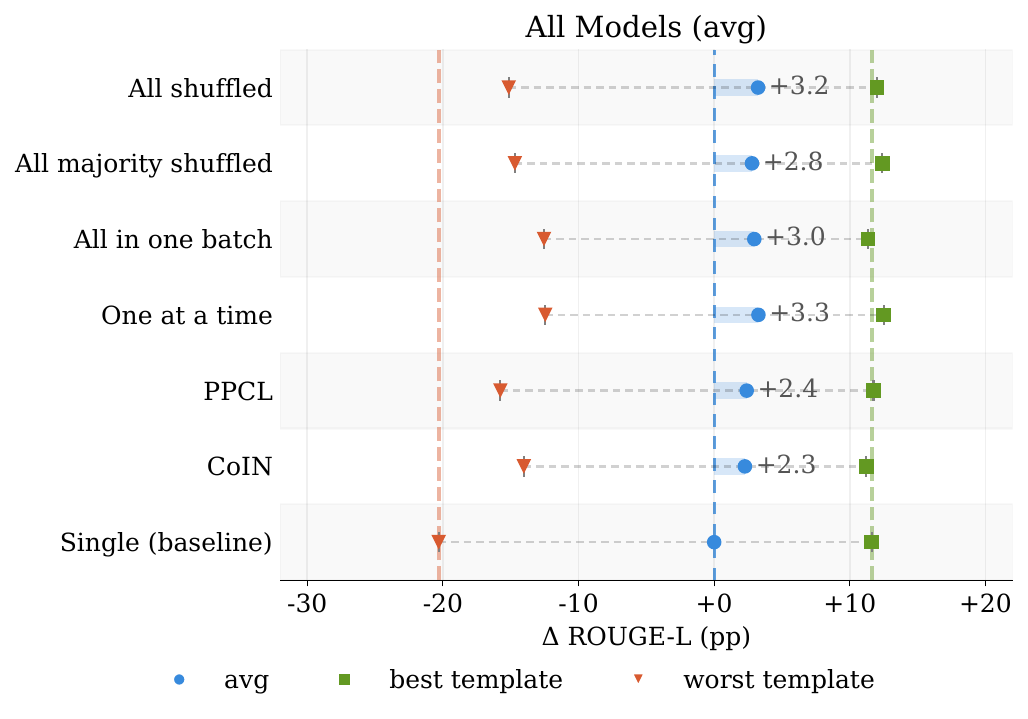}
    \caption{Performance of each method relative to IFT in percentage points averaged across models. The circle, square, and triangle mark the average, best, and worst template score respectively; the dashed span reflects sensitivity to prompt variance. Vertical lines extending from IFT serve as reference. A star above a marker indicates membership in the statistically best group.}
    \label{fig:robustness}
\end{figure}

\begin{table*}[htbp]
  \centering
  \caption{
    Raw Rouge-L scores on the test set.
    For each model we report the mean across all templates (\textit{avg}), the best-template score (\textit{best}), and the worst-template score (\textit{worst}). The upper block lists the two references that require no robustness-oriented training: the untuned base checkpoints and 4-shot in-context learning (\textsc{ICL}) on those
    same checkpoints. \textit{Single (5 ep.)} is the step-matched control, which trains the single-template baseline for five epochs to approximately match the optimizer-step count of the multi-template runs. It was run on the two small models only. \textbf{Bold} values belong to the statistically best group per column over the one-epoch trained methods, as in the main body of the paper.
  }
  \label{tab:rougeL_raw}
  \footnotesize
  \setlength{\tabcolsep}{4pt}
  \begin{tabular}{l ccc ccc ccc ccc}
    \toprule
    & \multicolumn{3}{c}{Qwen3-0.6B}
    & \multicolumn{3}{c}{Qwen3-8B}
    & \multicolumn{3}{c}{Llama3.2-1B}
    & \multicolumn{3}{c}{Llama3.1-8B} \\
    \cmidrule(lr){2-4}\cmidrule(lr){5-7}\cmidrule(lr){8-10}\cmidrule(lr){11-13}
    Method & avg & best & worst & avg & best & worst & avg & best & worst & avg & best & worst \\
    \midrule
    \multicolumn{13}{l}{\textit{No robustness training}} \\
    Base model            & 0.093 & 0.218 & 0.025 & 0.434 & 0.640 & 0.191 & 0.051 & 0.089 & 0.030 & 0.060 & 0.138 & 0.024 \\
    \textsc{ICL}        & 0.457 & 0.579 & 0.296 & 0.660 & 0.769 & 0.469 & 0.192 & 0.291 & 0.056 & 0.375 & 0.483 & 0.211 \\
    \midrule
    \multicolumn{13}{l}{\textit{Trained}} \\
    Single (baseline)     & 0.514 & 0.618 & 0.308 & 0.634 & 0.753 & 0.416 & 0.479 & \textbf{0.629} & 0.210 & 0.639 & \textbf{0.730} & \textbf{0.519} \\
    Single (5 ep.)        & 0.519 & 0.642 & 0.354 & --    & --    & --    & 0.494 & 0.607 & 0.242 & --    & --    & --    \\
    All shuffled          & 0.554 & 0.632 & 0.406 & 0.668 & \textbf{0.770} & \textbf{0.485} & \textbf{0.531} & 0.623 & 0.310 & 0.642 & 0.721 & 0.458 \\
    All majority shuffled & 0.549 & 0.635 & 0.416 & 0.663 & \textbf{0.769} & \textbf{0.477} & \textbf{0.531} & \textbf{0.635} & 0.311 & 0.634 & 0.722 & 0.473 \\
    All in one batch      & 0.556 & 0.627 & \textbf{0.452} & 0.655 & 0.758 & 0.467 & 0.524 & 0.608 & \textbf{0.379} & \textbf{0.648} & 0.726 & 0.464 \\
    One at a time         & \textbf{0.563} & \textbf{0.648} & \textbf{0.449} & 0.653 & 0.760 & 0.462 & \textbf{0.534} & \textbf{0.634} & \textbf{0.363} & \textbf{0.645} & 0.723 & 0.493 \\
    \textsc{PPCL}                  & 0.521 & 0.605 & 0.368 & 0.659 & 0.751 & \textbf{0.480} & \textbf{0.531} & \textbf{0.639} & 0.316 & \textbf{0.649} & \textbf{0.741} & 0.470 \\
    \textsc{CoIN}                  & 0.529 & 0.607 & 0.392 & \textbf{0.673} & \textbf{0.774} & \textbf{0.480} & 0.511 & 0.613 & 0.302 & 0.643 & 0.720 & \textbf{0.530} \\
    \bottomrule
  \end{tabular}
\end{table*}

\section{Validation Performance over Training}
\label{app:valcurves}

\autoref{tab:val_curves} reports validation Rouge-L at every checkpoint during Llama3.1-8B instruction fine-tuning. Across all methods, the largest performance improvement comes in the first 1{,}000 steps. No run improves by more than four points after 1,000 steps, except \textsc{One-at-a-Time}, which oscillates by up to ten points without a trend, and no run climbs monotonically. While the base model largely fails to produce the required output format, the checkpoints after 1{,}000 steps follow the instruction format. Still LoRA tuning contributes little additional benefit afterwards.

\begin{table*}[htbp]
  \centering
  \small
  \caption{
    Validation Rouge-L ($\times 100$) per checkpoint for Llama3.1-8B. Dashes mark checkpoints beyond the end of a run: single-template training covers fewer optimizer steps than multi-template training, and the majority-template runs fewer than the all-template runs. The final checkpoints close to the next thousand steps are rounded up.
  }\label{tab:val_curves}
  \begin{tabular}{@{}rccccccc@{}}
    \toprule
    \textbf{Step} & \textbf{Single} & \textbf{All shuffled} & \textbf{All maj.\ shuffled}
      & \textbf{One-at-a-Time} & \textbf{All-in-One-Batch} & \textbf{\textsc{PPCL}} & \textbf{\textsc{CoIN}} \\
    \midrule
           0 & 5.96  & 5.96  & 5.96  & 5.96  & 5.96  & 5.96  & 5.96  \\
     1{,}000 & 58.61 & 64.70 & 56.25 & 54.57 & 64.63 & 61.33 & 59.88 \\
     2{,}000 & 56.51 & 65.12 & 57.43 & 59.82 & 65.00 & 61.91 & 59.87 \\
     3{,}000 & --    & 65.63 & 58.01 & 59.44 & 64.92 & 58.80 & 59.08 \\
     4{,}000 & --    & 65.53 & 59.06 & 64.26 & 66.27 & 60.36 & 60.26 \\
     5{,}000 & --    & 67.75 & 59.21 & 56.50 & 65.50 & 60.60 & 60.75 \\
     6{,}000 & --    & 67.15 & 59.37 & 58.08 & 64.89 & 60.40 & 60.22 \\
     7{,}000 & --    & 67.20 & 58.82 & 57.41 & 65.06 & 61.45 & 61.28 \\
     8{,}000 & --    & 64.98 & 59.04 & 59.81 & 66.30 & 57.83 & 61.42 \\
     9{,}000 & --    & 66.06 & --    & 58.38 & 66.13 & --    & --    \\
    10{,}000 & --    & 66.69 & --    & 58.28 & 64.59 & --    & --    \\
    \bottomrule
  \end{tabular}
\end{table*}

\section{Rank Classification Accuracy}
\label{app:rankcls}

The T0-benchmark was originally evaluated using rank classification accuracy. Therefore, making the generative Rouge-L our primary metric is a deliberate choice. First, in rank classification accuracy the model is handed the candidate set and is asked only for a relative ordering of it, it is not used in a generative setting. Second, it only requires that the correct option remains marginally more likely, while the model's generative behavior could shift substantially across reformulations. Thus, a model can appear more prompt-robust under rank classification accuracy than under generative metrics. Third, constraining the model to a fixed candidate set compresses the differences between methods that our comparison depends on: on Qwen3-8B every method lands within $0.3$ accuracy points of every other (see \autoref{tab:rank_cls}), leaving almost no signal to separate them. Fourth, rank classification accuracy only works for the test split of our dataset because the training and validation splits incorporate examples without concrete answer sets.

\begin{table}[htbp]
  \centering
  \small
  \caption{
    Rank classification accuracy on the test split. \emph{avg} is the mean across all templates, \emph{best} and \emph{worst} the
    best- and worst-template scores.
  }\label{tab:rank_cls}
  \begin{tabular}{@{}lccc@{}}
    \toprule
    \textbf{Method} & \textbf{avg} & \textbf{best} & \textbf{worst} \\
    \midrule
    \multicolumn{4}{@{}l}{\textit{Llama3.2-1B}} \\
    \textsc{Single} (baseline) & 0.4821 & 0.5447 & 0.4025 \\
    \textsc{All shuffled}      & 0.4939 & 0.5564 & 0.3919 \\
    \textsc{All-in-One-Batch}  & 0.4706 & 0.5651 & 0.3500 \\
    \textsc{One-at-a-Time}     & 0.5043 & 0.5726 & 0.4178 \\
    \textsc{PPCL}              & 0.4799 & 0.5469 & 0.3837 \\
    \textsc{CoIN}              & 0.4835 & 0.5616 & 0.3969 \\
    \midrule
    \multicolumn{4}{@{}l}{\textit{Llama3.1-8B}} \\
    \textsc{Single} (baseline) & 0.5446 & 0.6514 & 0.4578 \\
    \textsc{All shuffled}      & 0.5644 & 0.6474 & 0.4555 \\
    \textsc{All-in-One-Batch}  & 0.5469 & 0.6293 & 0.4365 \\
    \textsc{One-at-a-Time}     & 0.5520 & 0.6557 & 0.4491 \\
    \textsc{PPCL}              & 0.5873 & 0.6785 & 0.4871 \\
    \textsc{CoIN}              & 0.5649 & 0.6929 & 0.4493 \\
    \midrule
    \multicolumn{4}{@{}l}{\textit{Qwen3-0.6B}} \\
    \textsc{Single} (baseline) & 0.4776 & 0.5314 & 0.4286 \\
    \textsc{All shuffled}      & 0.4849 & 0.5435 & 0.4309 \\
    \textsc{All-in-One-Batch}  & 0.4712 & 0.5092 & 0.4295 \\
    \textsc{One-at-a-Time}     & 0.4752 & 0.5195 & 0.4276 \\
    \textsc{PPCL}              & 0.4739 & 0.5287 & 0.4258 \\
    \textsc{CoIN}              & 0.4781 & 0.5256 & 0.4258 \\
    \midrule
    \multicolumn{4}{@{}l}{\textit{Qwen3-8B}} \\
    \textsc{Single} (baseline) & 0.4966 & 0.5295 & 0.4568 \\
    \textsc{All shuffled}      & 0.4979 & 0.5298 & 0.4597 \\
    \textsc{All-in-One-Batch}  & 0.4963 & 0.5295 & 0.4579 \\
    \textsc{One-at-a-Time}     & 0.4982 & 0.5296 & 0.4619 \\
    \textsc{PPCL}              & 0.4956 & 0.5275 & 0.4580 \\
    \textsc{CoIN}              & 0.4979 & 0.5324 & 0.4603 \\
    \bottomrule
  \end{tabular}
\end{table}

\section{Why \textsc{PPCL} and \textsc{CoIN} Fail}
\label{app:diagnostics}

Both auxiliary objectives fail to reliably improve on the average and worst-case performance compared to the far cheaper data-construction schedules. To determine whether they fail to optimize their objective or fail to generalize from it, we measure the quantity each loss penalizes directly, after training.

\paragraph{Measured quantities}
For \textsc{CoIN} we compute the contrastive difference between the same-example cross-template similarity and the average cross-example similarity over the model's hidden states at a given readout position. This mirrors the contrastive target of \textsc{CoIN} without requiring its training-pair construction at evaluation time. A higher Diff indicates stronger semantic-over-lexical alignment, that is, exactly what \textsc{CoIN} optimizes for. For the label readout we check the hidden states of the last label token and for the prompt readout the last prompt token. The results are in \autoref{tab:coin_diag}.

For \textsc{PPCL} we compute the cross-template Jensen--Shannon divergence of the output distributions exactly matching its training objective. A lower JS corresponds to more consistent generation across templates. Again we read out either the average of all label tokens, as in the \textsc{PPCL} paper, or from the last prompt token. The results are in \autoref{tab:ppcl_diag}. 

\begin{figure*}[tp]
    \centering
    \includegraphics[width=\textwidth]{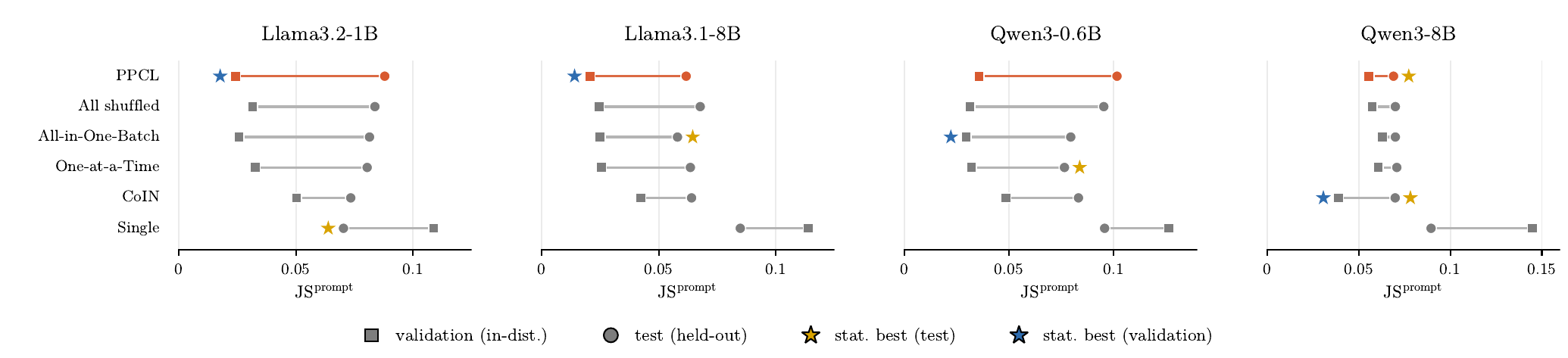}
    \caption{The cross-template Jensen--Shannon divergence of the output distributions on the \texttt{prompt} readout. Squares show the score on the validation split, while circles show the score on the test split. A star next to a marker indicates membership in the statistically best group for that readout. \textsc{PPCL} achieves for the Llama models substantially lower divergence in the validation set, but this advantage disappears in the held-out test set.}
    \label{fig:jsdiv_per_model}
\end{figure*}

\begin{table}[htbp]
  \centering
  \small
  \setlength{\tabcolsep}{4pt}
  \caption{
    \textsc{CoIN} diagnostic. Diff (higher = stronger semantic-over-lexical alignment) at the
    prompt ($p$) and label ($l$) readouts, on the held-out test split and the in-distribution
    validation split. \textsc{CoIN} dominates at the label readout, where its loss is applied,
    but not at the prompt readout, where generation begins.
  }\label{tab:coin_diag}
  \begin{tabular}{@{}lcccc@{}}
    \toprule
    & \multicolumn{2}{c}{\textbf{test}} & \multicolumn{2}{c}{\textbf{val}} \\
    \cmidrule(lr){2-3}\cmidrule(lr){4-5}
    \textbf{Method} & Diff$^{p}$ & Diff$^{l}$ & Diff$^{p}$ & Diff$^{l}$ \\
    \midrule
    \multicolumn{5}{@{}l}{\textit{Llama3.2-1B}} \\
    \textsc{Single} (baseline) & 0.0861 & 0.0107 & 0.2051 & 0.0147 \\
    \textsc{All shuffled}      & 0.2417 & 0.0156 & 0.4265 & 0.0186 \\
    \textsc{All-in-One-Batch}  & 0.1848 & 0.0132 & 0.3320 & 0.0174 \\
    \textsc{One-at-a-Time}     & 0.2603 & 0.0134 & 0.4141 & 0.0165 \\
    \textsc{PPCL}              & 0.1169 & 0.0119 & 0.2542 & 0.0154 \\
    \textsc{CoIN}              & 0.2095 & 0.7792 & 0.3684 & 0.7907 \\
    \midrule
    \multicolumn{5}{@{}l}{\textit{Llama3.1-8B}} \\
    \textsc{Single} (baseline) & 0.1081 & 0.0033 & 0.2405 & 0.0048 \\
    \textsc{All shuffled}      & 0.1324 & 0.0031 & 0.3149 & 0.0045 \\
    \textsc{All-in-One-Batch}  & 0.1068 & 0.0030 & 0.2634 & 0.0043 \\
    \textsc{One-at-a-Time}     & 0.1223 & 0.0031 & 0.2944 & 0.0046 \\
    \textsc{PPCL}              & 0.1209 & 0.0029 & 0.2738 & 0.0042 \\
    \textsc{CoIN}              & 0.1234 & 0.7896 & 0.2843 & 0.8064 \\
    \midrule
    \multicolumn{5}{@{}l}{\textit{Qwen3-0.6B}} \\
    \textsc{Single} (baseline) & 0.0506 & 0.0024 & 0.1027 & 0.0035 \\
    \textsc{All shuffled}      & 0.0633 & 0.0016 & 0.1351 & 0.0024 \\
    \textsc{All-in-One-Batch}  & 0.0594 & 0.0013 & 0.1283 & 0.0019 \\
    \textsc{One-at-a-Time}     & 0.0555 & 0.0019 & 0.1267 & 0.0028 \\
    \textsc{PPCL}              & 0.0538 & 0.0024 & 0.1070 & 0.0037 \\
    \textsc{CoIN}              & 0.0764 & 0.7572 & 0.1622 & 0.7924 \\
    \midrule
    \multicolumn{5}{@{}l}{\textit{Qwen3-8B}} \\
    \textsc{Single} (baseline) & 0.0208 & 0.0627 & 0.0298 & 0.0581 \\
    \textsc{All shuffled}      & 0.0214 & 0.0244 & 0.0331 & 0.0259 \\
    \textsc{All-in-One-Batch}  & 0.0202 & 0.0178 & 0.0311 & 0.0194 \\
    \textsc{One-at-a-Time}     & 0.0222 & 0.0269 & 0.0339 & 0.0294 \\
    \textsc{PPCL}              & 0.0209 & 0.0146 & 0.0321 & 0.0157 \\
    \textsc{CoIN}              & 0.0514 & 0.7355 & 0.0582 & 0.7351 \\
    \bottomrule
  \end{tabular}
\end{table}

\begin{table}[htbp]
  \centering
  \small
  \setlength{\tabcolsep}{4pt}
  \caption{
    \textsc{PPCL} diagnostic. Cross-template Jensen--Shannon divergence (lower = more consistent
    generation) at the prompt ($p$) and label ($l$) readouts. \textsc{PPCL} reaches the lowest
    validation JS on the Llama models, so its objective succeeds in-distribution, but the
    advantage does not transfer to the held-out test split, where it is the worst multi-template
    method on Llama3.2-1B and Qwen3-0.6B.
  }\label{tab:ppcl_diag}
  \begin{tabular}{@{}lcccc@{}}
    \toprule
    & \multicolumn{2}{c}{\textbf{test}} & \multicolumn{2}{c}{\textbf{val}} \\
    \cmidrule(lr){2-3}\cmidrule(lr){4-5}
    \textbf{Method} & JS$^{p}$ & JS$^{l}$ & JS$^{p}$ & JS$^{l}$ \\
    \midrule
    \multicolumn{5}{@{}l}{\textit{Llama3.2-1B}} \\
    \textsc{Single} (baseline) & 0.0703 & 0.0325 & 0.1088 & 0.0466 \\
    \textsc{All shuffled}      & 0.0837 & 0.0375 & 0.0315 & 0.0145 \\
    \textsc{All-in-One-Batch}  & 0.0814 & 0.0361 & 0.0257 & 0.0121 \\
    \textsc{One-at-a-Time}     & 0.0804 & 0.0344 & 0.0326 & 0.0148 \\
    \textsc{PPCL}              & 0.0879 & 0.0400 & 0.0242 & 0.0112 \\
    \textsc{CoIN}              & 0.0734 & 0.0307 & 0.0503 & 0.0240 \\
    \midrule
    \multicolumn{5}{@{}l}{\textit{Llama3.1-8B}} \\
    \textsc{Single} (baseline) & 0.0848 & 0.0379 & 0.1139 & 0.0450 \\
    \textsc{All shuffled}      & 0.0677 & 0.0301 & 0.0245 & 0.0112 \\
    \textsc{All-in-One-Batch}  & 0.0580 & 0.0258 & 0.0250 & 0.0111 \\
    \textsc{One-at-a-Time}     & 0.0635 & 0.0278 & 0.0256 & 0.0115 \\
    \textsc{PPCL}              & 0.0617 & 0.0283 & 0.0206 & 0.0095 \\
    \textsc{CoIN}              & 0.0640 & 0.0286 & 0.0423 & 0.0198 \\
    \midrule
    \multicolumn{5}{@{}l}{\textit{Qwen3-0.6B}} \\
    \textsc{Single} (baseline) & 0.0957 & 0.0856 & 0.1264 & 0.0557 \\
    \textsc{All shuffled}      & 0.0953 & 0.0770 & 0.0313 & 0.0144 \\
    \textsc{All-in-One-Batch}  & 0.0795 & 0.0660 & 0.0295 & 0.0135 \\
    \textsc{One-at-a-Time}     & 0.0765 & 0.0650 & 0.0321 & 0.0149 \\
    \textsc{PPCL}              & 0.1016 & 0.0913 & 0.0357 & 0.0167 \\
    \textsc{CoIN}              & 0.0832 & 0.0709 & 0.0486 & 0.0227 \\
    \midrule
    \multicolumn{5}{@{}l}{\textit{Qwen3-8B}} \\
    \textsc{Single} (baseline) & 0.0895 & 0.0814 & 0.1449 & 0.0739 \\
    \textsc{All shuffled}      & 0.0700 & 0.0619 & 0.0574 & 0.0278 \\
    \textsc{All-in-One-Batch}  & 0.0700 & 0.0623 & 0.0628 & 0.0281 \\
    \textsc{One-at-a-Time}     & 0.0708 & 0.0630 & 0.0606 & 0.0299 \\
    \textsc{PPCL}              & 0.0690 & 0.0616 & 0.0556 & 0.0245 \\
    \textsc{CoIN}              & 0.0699 & 0.0632 & 0.0390 & 0.0170 \\
    \bottomrule
  \end{tabular}
\end{table}

\section{Gradient Interference}
\label{app:interference}

\paragraph{Gradient Interference measurements}
During \textsc{All-in-One-Batch} training we compute per-template gradients and summarize their disagreement with four scale-invariant statistics:
\begin{itemize}
  \item \textbf{Mean pairwise cosine similarity} over all template-gradient pairs ($1$~=~collinear, $0$~=~orthogonal, negative~=~opposed), i.e.\ whether the templates share a common descent direction.
  \item \textbf{Coordinate-level sign conflict}: the fraction of parameters on which at least two template gradients disagree in sign ($0$~=~every parameter agrees, $1$~=~every parameter contested).
  \item \textbf{Magnitude cancellation}: the fraction of the average per-template gradient norm that the naive gradient average fails to retain ($0$~=~averaging keeps the full norm, $1$~=~gradients cancel entirely).
  \item \textbf{Entropy-based effective rank}: how many distinct directions the per-template gradients span, from $1$ (one shared direction) up to the number of templates $T$ \citep{roy2007effective}.
\end{itemize}

We compute these metrics every 1{,}000 steps for both Llama models (see \autoref{tab:interference}). In addition, we report the per-layer pairwise cosine similarity and sign conflict for these models, see \autoref{tab:layer_interference_1b} and \autoref{tab:layer_interference_8b}.

\paragraph{Results}
Per-template gradients agree only coarsely, at a mean cosine similarity of about $0.54$ on both models, and the disagreement is pronounced at the coordinate level: templates push in opposite directions on 57\% (Llama3.2-1B) and 64\% (Llama3.1-8B) of the trainable parameters, roughly 18--20\% of the gradient magnitude cancels under averaging, and the per-template gradients span about three effective directions rather than one. These values are stable across training (\autoref{tab:interference}). Interference decreases with depth (\autoref{tab:layer_interference_1b}, \autoref{tab:layer_interference_8b}): sign conflict falls by roughly 23 points from the first to the last block of Llama3.1-8B, which is consistent with the conflict originating in the surface form of the prompt and dissipating as representations abstract away from wording. We observe no meaningful difference between the attention and feed-forward layers. Mixing templates within a batch forces the optimizer to reconcile several competing directions in a single update, whereas the template-homogeneous batches of \textsc{One-at-a-Time} never co-locate those conflicting gradients.

\begin{table*}[htbp]
  \centering
  \small
  \caption{
    Interference statistics over training for \textsc{All-in-One-Batch}. \emph{Avg.} is the mean over all checkpoints for Llama3.2-1B. For Llama3.1-8B it excludes $t=0$, which is an outlier due to the LoRA adapters being zero-initialized, and the measurement preceding any training.
  }\label{tab:interference}
  \begin{tabular}{@{}lcccccccc@{}}
    \toprule
    \textbf{Statistic} & $t=0$ & 1001 & 2002 & 3003 & 4004 & 5005 & 6006 & \textbf{Avg.} \\
    \midrule
    \multicolumn{9}{@{}l}{\textit{Llama3.2-1B}} \\
    cos           & 0.531 & 0.532 & 0.541 & 0.546 & 0.550 & 0.551 & 0.549 & \textbf{0.543} \\
    cos (min)     & 0.289 & 0.294 & 0.293 & 0.292 & 0.299 & 0.295 & 0.295 & \textbf{0.294} \\
    sign-conflict & 0.615 & 0.571 & 0.564 & 0.565 & 0.566 & 0.560 & 0.561 & \textbf{0.572} \\
    cancellation  & 0.217 & 0.204 & 0.202 & 0.194 & 0.189 & 0.185 & 0.191 & \textbf{0.197} \\
    eff-rank      & 3.273 & 2.988 & 2.852 & 2.836 & 2.802 & 2.805 & 2.810 & \textbf{2.909} \\
    \midrule
    \multicolumn{9}{@{}l}{\textit{Llama3.1-8B}} \\
    cos           & 0.502 & 0.514 & 0.564 & 0.567 & 0.556 & 0.551 & 0.544 & \textbf{0.549} \\
    cos (min)     & 0.274 & 0.251 & 0.313 & 0.311 & 0.295 & 0.272 & 0.272 & \textbf{0.286} \\
    sign-conflict & 0.379 & 0.667 & 0.640 & 0.634 & 0.634 & 0.629 & 0.630 & \textbf{0.639} \\
    cancellation  & 0.232 & 0.204 & 0.169 & 0.167 & 0.161 & 0.169 & 0.181 & \textbf{0.175} \\
    eff-rank      & 3.505 & 3.148 & 2.860 & 2.815 & 2.974 & 2.854 & 2.985 & \textbf{2.939} \\
    \bottomrule
  \end{tabular}
\end{table*}

\begin{table}[htbp]
  \centering
  \small
  \caption{
    Per-layer cosine similarity and sign conflict for the attention and feed-forward sublayers of Llama3.2-1B, averaged over training. Interference decreases with depth.
  }\label{tab:layer_interference_1b}
  \begin{tabular}{@{}rcccc@{}}
    \toprule
    & \multicolumn{2}{c}{Attention} & \multicolumn{2}{c}{Feed-forward} \\
    \cmidrule(lr){2-3}\cmidrule(lr){4-5}
    \textbf{Layer} & cos & sign-conf. & cos & sign-conf. \\
    \midrule
     0 & 0.483 & 0.714 & 0.499 & 0.687 \\
     1 & 0.485 & 0.710 & 0.547 & 0.692 \\
     2 & 0.480 & 0.715 & 0.484 & 0.694 \\
     3 & 0.495 & 0.704 & 0.491 & 0.690 \\
     4 & 0.502 & 0.703 & 0.500 & 0.685 \\
     5 & 0.512 & 0.692 & 0.507 & 0.681 \\
     6 & 0.526 & 0.683 & 0.512 & 0.680 \\
     7 & 0.536 & 0.683 & 0.519 & 0.676 \\
     8 & 0.543 & 0.675 & 0.534 & 0.664 \\
     9 & 0.561 & 0.662 & 0.550 & 0.644 \\
    10 & 0.566 & 0.648 & 0.560 & 0.628 \\
    11 & 0.572 & 0.635 & 0.583 & 0.610 \\
    12 & 0.587 & 0.620 & 0.591 & 0.601 \\
    13 & 0.609 & 0.609 & 0.600 & 0.587 \\
    14 & 0.605 & 0.604 & 0.627 & 0.568 \\
    15 & 0.626 & 0.579 & 0.646 & 0.543 \\
    \bottomrule
  \end{tabular}
\end{table}

\begin{table*}[htbp]
  \centering
  \small
  \caption{
    Per-layer cosine similarity and sign conflict for the attention and feed-forward sublayers of Llama3.1-8B, averaged over training. Layers 0--15 are shown on the left and 16--31 on the right. Interference decreases with depth.
  }\label{tab:layer_interference_8b}
  \begin{tabular}{@{}rcccc@{\hspace{2em}}rcccc@{}}
    \toprule
    & \multicolumn{2}{c}{Attention} & \multicolumn{2}{c}{Feed-forward}
    & & \multicolumn{2}{c}{Attention} & \multicolumn{2}{c}{Feed-forward} \\
    \cmidrule(lr){2-3}\cmidrule(lr){4-5}\cmidrule(lr){7-8}\cmidrule(lr){9-10}
    \textbf{Layer} & cos & sign-conf. & cos & sign-conf.
      & \textbf{Layer} & cos & sign-conf. & cos & sign-conf. \\
    \midrule
     0 & 0.347 & 0.811 & 0.336 & 0.780 & 16 & 0.519 & 0.661 & 0.543 & 0.651 \\
     1 & 0.343 & 0.783 & 0.450 & 0.767 & 17 & 0.528 & 0.658 & 0.539 & 0.644 \\
     2 & 0.353 & 0.788 & 0.374 & 0.765 & 18 & 0.510 & 0.655 & 0.536 & 0.642 \\
     3 & 0.365 & 0.776 & 0.368 & 0.767 & 19 & 0.511 & 0.655 & 0.540 & 0.634 \\
     4 & 0.382 & 0.776 & 0.371 & 0.764 & 20 & 0.529 & 0.641 & 0.540 & 0.631 \\
     5 & 0.381 & 0.767 & 0.383 & 0.757 & 21 & 0.539 & 0.640 & 0.544 & 0.628 \\
     6 & 0.422 & 0.748 & 0.404 & 0.748 & 22 & 0.524 & 0.640 & 0.549 & 0.623 \\
     7 & 0.449 & 0.733 & 0.438 & 0.730 & 23 & 0.537 & 0.633 & 0.546 & 0.620 \\
     8 & 0.465 & 0.718 & 0.454 & 0.719 & 24 & 0.536 & 0.634 & 0.551 & 0.622 \\
     9 & 0.475 & 0.715 & 0.468 & 0.708 & 25 & 0.539 & 0.629 & 0.550 & 0.616 \\
    10 & 0.490 & 0.700 & 0.495 & 0.693 & 26 & 0.558 & 0.616 & 0.557 & 0.611 \\
    11 & 0.511 & 0.683 & 0.513 & 0.684 & 27 & 0.559 & 0.611 & 0.570 & 0.603 \\
    12 & 0.507 & 0.684 & 0.525 & 0.679 & 28 & 0.566 & 0.605 & 0.575 & 0.595 \\
    13 & 0.521 & 0.676 & 0.538 & 0.667 & 29 & 0.580 & 0.593 & 0.589 & 0.586 \\
    14 & 0.528 & 0.668 & 0.539 & 0.663 & 30 & 0.566 & 0.595 & 0.603 & 0.574 \\
    15 & 0.523 & 0.666 & 0.550 & 0.656 & 31 & 0.600 & 0.573 & 0.616 & 0.550 \\
    \bottomrule
  \end{tabular}
\end{table*}

\section{Statement on AI use and Risks}
The authors used AI-assisted writing tools for language editing. All content was verified by the authors.

The authors do not see direct risks from publishing the paper as all used data is from well-known benchmarks used before.

\end{document}